\documentclass[10pt,twocolumn,letterpaper]{article}
\usepackage[pagenumbers]{wacv} % arXiv version
\def\wacvPaperID{2361}
\def\confName{WACV}
\def\confYear{2025}

\usepackage{papermacros}
\usepackage{interval}
\usepackage{standalone}
\usepackage[bookmarks=false]{hyperref}
\usepackage[capitalize]{cleveref}
\crefname{section}{Sec.}{Secs.}
\Crefname{section}{Section}{Sections}
\Crefname{table}{Table}{Tables}
\crefname{table}{Tab.}{Tabs.}

\begin{document}
\title{%
Balancing Shared and Task-Specific Representations: \\A Hybrid Approach to Depth-Aware Video Panoptic Segmentation%
}

\author{Kurt H.W. Stolle\\
Eindhoven University of Technology\\The Netherlands\\
{\tt\small k.h.w.stolle@tue.nl}
}
\maketitle

% ========== A B S T R A C T ==========
\begin{abstract}
In this work, we present Multiformer, a novel approach to depth-aware video panoptic segmentation (DVPS) based on the mask transformer paradigm. 
Our method learns object representations that are shared across segmentation, monocular depth estimation, and object tracking subtasks. 
In contrast to recent unified approaches that progressively refine a common object representation, we propose a hybrid method using task-specific branches within each decoder block, ultimately fusing them into a shared representation at the block interfaces.
Extensive experiments on the Cityscapes-DVPS and SemKITTI-DVPS datasets demonstrate that Multiformer achieves state-of-the-art performance across all DVPS metrics, outperforming previous methods by substantial margins. 
With a ResNet-50 backbone, Multiformer surpasses the previous best result by 3.0~DVPQ points while also improving depth estimation accuracy. 
Using a Swin-B backbone, Multiformer further improves performance by 4.0~DVPQ points.
Multiformer also provides valuable insights into the design of multi-task decoder architectures. 
\end{abstract}

% ========== I N T R O D U C T I O N ==========
\section{Introduction}
\label{sec:intro}

% ---------- Hero figure ------------
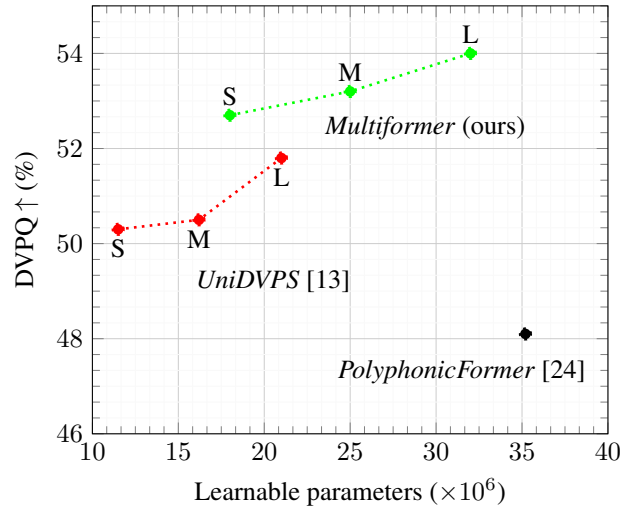
\begin{figure}[!t]
    \centering
    \begin{adjustbox}{max width=\linewidth}
        \centering
\begin{tikzpicture}
\begin{axis}[
    xlabel={Learnable parameters ($\times 10^6$)},
    ylabel={DVPQ $\uparrow$ (\%)},
    xmin=10, xmax=40,
    ymin=46, ymax=55,
    minor tick num=5,
    grid=both,
    grid style={line width=.1pt, draw=gray!5},
    major grid style={line width=.2pt,draw=gray!40},
    scatter src=explicit symbolic,
    scatter/classes={
        Polyphonic={mark=square*},
        UniDVPS={mark=*},
        Ours={mark=triangle*}
    },
    scatter/position=relative,
    nodes near coords,
    nodes near coords align={horizontal},
    every axis plot/.append style={line width=1pt},
    every node near coord/.append style={
        text=black, 
        text centered,
    },
    point meta=explicit symbolic,
    label/.style={
        text=black,
        text centered,
        inner sep=0pt,
    },
]
% PolyphonicFormer
\pgfplotstableread{
NAME                  PARAMS      DVPQ 
{L}    35.2        48.1
}\dataPoly
\addplot[
    scatter,
    dotted,
    black
] table[
    x=PARAMS,
    y=DVPQ,
] {\dataPoly};
\node[anchor=center, label, xshift=-8mm, yshift=2mm] at (axis cs:35.0,47.0) {\textit{PolyphonicFormer}~\protect\cite{Yuan2021PolyphonicFormer}};

% UniDVPS
\pgfplotstableread{
NAME                  PARAMS      DVPQ 
{S}         11.5        50.3
{M}         16.2        50.5
{L}         21.0        51.8
}\dataUniDVPS
\addplot[
    scatter,
    dotted,
    red,
    every node near coord/.append style={
        anchor=north,
    },
] table[
    x=PARAMS,
    y=DVPQ,
    meta=NAME,
] {\dataUniDVPS};
\node[anchor=west, label, yshift=2mm] at (axis cs:15.5,48.9) {\textit{UniDVPS}~\protect\cite{JiYeon2024}};

% Ours
\pgfplotstableread{
NAME                  PARAMS      DVPQ 
{S}     18.0        52.7
{M}     25.0        53.2
{L}     32.0        54.0
}\dataOurs
\addplot[
    scatter,
    dotted,
    green,
] table[
    x=PARAMS,
    y=DVPQ,
    meta=NAME,
] {\dataOurs};

\node[anchor=north, label, xshift=10mm, yshift=-2mm] at (axis cs:25.0,53.2) {\textit{Multiformer}~(ours)};
\end{axis}
\end{tikzpicture}
    \end{adjustbox}
    \caption{
        \textbf{Model size vs. Depth-aware Video Panoptic Quality.}
        Evaluated on Cityscapes-DVPS with ResNet-50 as the backbone.%      
    }
    \label{fig:time}
\end{figure}

The integration of geometric perception and semantic understanding is crucial for advanced computer vision applications.
Depth-aware video panoptic segmentation (DVPS)~\cite{Qiao2020ViP} has emerged as a challenging task that combines monocular depth estimation, object tracking and segmentation, offering a comprehensive solution for 3D scene understanding from a single camera.

Researchers who address the DVPS task through a unified network have found that combining semantic and geometric embeddings leads to both improved DVPS and subtask quality. 
Recent DVPS approaches concentrate on either interactions between separate depth and segmentation representations~\cite{Qiao2020ViP,Petrovai2023MonoDVPS}, or propose fully shared representations~\cite{JiYeon2024}. 
While shared approaches offer benefits like smaller models and implicit multi-task learning, they may limit the degree to which task nuances can be captured by the model.

This work, called \textit{Multiformer}, balances these approaches, combining shared representation with task-specific modeling.
The key innovation lies in the novel decoder architecture, which learns a multi-task representation that is split into task-specific branches within each decoder block, but then combines these at the interfaces between decoder blocks.
This hybrid approach enables task-specific deep supervision of intra-decoder representations,  while also maintaining the benefits of shared representations.

A contribution of this work is comparing the \textit{Multiformer} design against a comprehensive space of alternative decoder designs.
This provides valuable insights into balancing task-specific and shared representations in multi-task vision models.
By striking a balance between task-specific and shared representations, \textit{Multiformer} achieves state-of-the-art performance in depth-aware video panoptic segmentation and its component tasks, as shown in \cref{fig:time}.
The main contributions of this work are as follows.
\begin{itemize}
    \item \textit{Multiformer}, a state-of-the-art DVPS model that balances shared and task-specific representations.
    \item An exploration of alternative decoder designs, including reimplementation of state-of-the-art methods.
\end{itemize}
% TODO: maybe move to abstract or keep in intro?
The \textit{Multiformer} code and trained models are available at \mbox{\tt
\href{https://research.khws.io/multiformer}{research.khws.io/multiformer}
}

% ========== R E L A T E D   W O R K ===========
\section{Related work}
\label{sec:related}

% ---------- DVPS ----------
\subsection{Depth-aware video panoptic segmentation}
\label{sec:related-work}

Depth-aware video panoptic segmentation (DVPS)~\cite{Qiao2020ViP} is the combined task of segmentation, depth estimation and object tracking. 
% TODO: check whether these are actually still the _only_ approaches
Currently, the following approaches have been proposed.

\noindent
\textbf{ViP-DeepLab~\cite{Qiao2020ViP}} first introduced the DVPS task, extending \textit{Panoptic-DeepLab}~\cite{Cheng2020Panoptic} with depth-aware video processing capabilities. 
The method employs a shared backbone architecture for feature extraction, complemented by task-specific CNN-based decoder heads dedicated to depth estimation, panoptic segmentation, and instance tracking.

\noindent
\textbf{MonoDVPS~\cite{Petrovai2023MonoDVPS}} enhances \textit{ViP-DeepLab}~\cite{Qiao2020ViP} by integrating semi-supervised components, thereby mitigating reliance on expensive ground-truth annotations. 
The method extends several semi-supervised approaches that have proven effective in monocular depth estimation~\cite{Godard2019Digging} to video panoptic segmentation.

\noindent
\textbf{PolyphonicFormer~\cite{Yuan2021PolyphonicFormer}} aims to unify the task-specific processing branches through `query reasoning' to enhance depth and tracking subtasks with instance-level semantic information.
The method uses a decoder based on \textit{Video K-Net}~\cite{Li2022Video} to learn how to reason about the interdependencies between separate task representations.
Although their method shares similarities with our decoder, the proposed method is characterized by the use of a shared representation, in contrast to using multiple task-specific features. 
In particular, the shared representations in the \textit{Multiformer} already embed all subtasks, while `query reasoning' facilitates the exchange of information between task-specific representations.

\noindent
\textbf{UniDVPS~\cite{JiYeon2024}} is a state-of-the-art DVPS model that adheres to the paradigm of unified object-level embeddings for multiple tasks. 
It proposes a query decoder architecture based on \textit{DETR}~\cite{Carion2020End}, where inter-task information exchange is learned in the network itself, rather than imposed through multiple task-specific decoders.
This entails using a common embedding for all subtasks, significantly reducing the amount of trainable parameters, and improving the efficiency of the network.
% TODO wording
While \textit{UniDVPS}~\cite{JiYeon2024} demonstrates the effectiveness of a fully shared approach, this work explores the balance between shared and task-specific embeddings. 
This balance enables the \textit{Multiformer} to capture task-specific nuances while maintaining a unified representation at the interface between decoder blocks.

% ---------- Mask2Former ----------
\subsection{Mask transformer}
\label{sec:related-mask}
Mask transformers~\cite{Cheng2021MaskFormer} represent an innovative class of models that leverage a transformer-based architecture to integrate object detection and segmentation tasks within a single framework.
The fundamental principle of mask transformers lies in the ability of the network to learn object-level representations by tailoring a set of learnable queries to the visual content depicted in the scene.
This capability is facilitated by a query decoder that sequentially applies cross-attention of these queries to the visual features.
Each object representation is then used for classification and combined with dense visual features to generate segmentation masks.
Recent advances introduced by \textit{Mask2Former}~\cite{Cheng2022Masked} enhance the query decoder through a masked-attention mechanism.
This masked-attention mechanism is a variation on cross-attention that ensures queries only focus on a specific region of the image features.
By generating segmentation masks after each decoder block, subsequent blocks can be focused to attend only to this region of interest, gradually refining the masks and queries' representations.
Moreover, this iterative approach enables \textit{deep supervision} of the queries, where the losses can be applied to the task-specific representations generated in each of the decoder blocks.
This approach has been shown to improve the convergence of the network as well as the segmentation quality~\cite{Cheng2022Masked}.

Currently, mask transformers have been implemented in a set of dense video computer vision tasks~\cite{Cheng2022Masked, Weng2023mask, Kim2022TubeFormer}, demonstrating consistent performance improvements over alternative approaches.
Although existing methods have adopted transformer-based architectures for DVPS~\cite{Yuan2021PolyphonicFormer, Cheng2022Masked}, the advantages of employing a mask transformer remain insufficiently investigated.

% ========== M E T H O D S ==========
\section{Method}
\label{sec:method}

% ---------- OVERVIEW ----------
%auto-ignore

% -- COLORS ------------------------------------------
\definecolor{NNQueryColor1}{HTML}{05D6ED}
\definecolor{NNQueryColor2}{HTML}{D94032} 
\definecolor{NNQueryColor3}{HTML}{F2E41D}
\definecolor{NNQueryColor4}{HTML}{CAF21D}
\definecolor{QueryColor2}{HTML}{05D6ED}
\definecolor{QueryColor3}{HTML}{D94032} 
\definecolor{QueryColor4}{HTML}{F2E41D}
\definecolor{QueryColor1}{HTML}{CAF21D}
\definecolor{MaskColor1}{HTML}{FA5B00}
\definecolor{MaskColor2}{HTML}{CF6223} 
\definecolor{MaskColor3}{HTML}{BC6D00}
\definecolor{MaskColor4}{HTML}{A55F37}
\definecolor{DepthColor1}{HTML}{00FAF9}
\definecolor{DepthColor2}{HTML}{28C7C6} 
\definecolor{DepthColor3}{HTML}{3B9493}
\definecolor{DepthColor4}{HTML}{3A6161}
% Other colors
\definecolor{ModuleFillColor}{HTML}{FAF9F7}
\definecolor{LayerFillColor}{HTML}{FFFEFC}
\definecolor{DrawColor}{HTML}{403E3C}
\definecolor{OpFillColor}{HTML}{FFFFFF}
\definecolor{FeatureFillColor}{HTML}{DAE1ED}
\definecolor{FeatureDrawColor}{HTML}{6F7279}
% Region colors
\definecolor{PastelColor1}{HTML}{E0F5FF}
\definecolor{PastelColor2}{HTML}{FCE0E1} 
\definecolor{PastelColor3}{HTML}{CCE7CF}
% \definecolor{QueryColor4}{HTML}{CAF21D}

\begin{figure*}[t]
    \centering
    % \begin{adjustbox}{max width=\textwidth, max height=20cm}
    % \centering
    % \includestandalone[mode=buildnew]{figure_overview}
    % \end{adjustbox}
    \includegraphics[width=\linewidth]{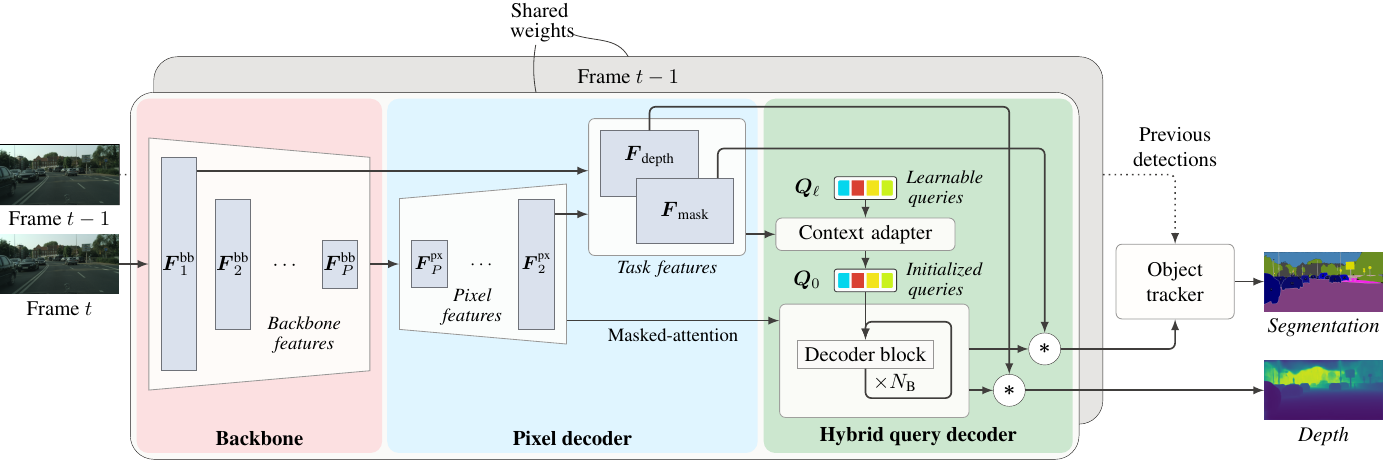}
    \caption{\textbf{Network overview.} \textit{Multiformer} is composed of a feature extraction \colorbox{PastelColor2}{backbone}, multi-scale \colorbox{PastelColor1}{pixel decoder},  \colorbox{PastelColor3}{hybrid query decoder}, and an object tracking module. Images are processed frame-by-frame, and the network outputs temporally consistent panoptic segmentation and depth.}
    \label{fig:architecture}
\end{figure*}

This section presents \textit{Multiformer}, a multi-task mask transformer model designed for simultaneous depth estimation and segmentation in video data.
A robust baseline is established through the replication of a state-of-the-art model employing the shared representation approach, which is reimplemented within the mask transformer~\cite{Cheng2022Masked} paradigm.
Subsequently, an innovative class of \textit{hybrid query decoders} is introduced. 

% --------- BASELINE NETWORK ----------
\subsection{Unified baseline network}
\label{sec:method-baseline}
Motivated by the recent success of the mask transformer paradigm in dense computer vision tasks~\cite{Cheng2022Masked, Weng2023mask, Kim2022TubeFormer}, this paper adopts and extends \textit{Mask2Former}~\cite{Cheng2022Masked}, a state-of-the-art universal segmentation architecture, to incorporate depth-aware video segmentation capabilities. 
To achieve this, the methods proposed in \textit{UniDVPS}~\cite{JiYeon2024} are followed to provide the aforementioned functionality.

\paragraph{Backbone.}
The video inputs are passed frame-by-frame to a pre-trained feature extractor~\cite{He2015Deep,Liu2021Swin}.
This `backbone' generates $P$ features that serve as input to subsequent components. 
The multi-scale \textit{backbone features} are denoted as $\mat{F}^\text{bb}_p$ for the feature level $p \in \{ 1 \cdots P \}$. 
Each $p$-th backbone feature has dimensions $C^\text{bb}_{p} \times H / 2^p \times W / 2^p$, where $H$ and $W$ represent the height and width of the input image, respectively, and $C^\text{bb}_p$ denotes the number of channels. 

\paragraph{Pixel decoder.}
The pixel decoder employs \textit{Multi-scale Deformable Attention}~\cite{Zhu2020Deformable} to produce $P-1$ features from all backbone features except the one with the highest resolution.
These \textit{pixel features} are expressed as $\matF^\text{px}_m$ at the level $m \in \{2\cdots P\}$.
All pixel features possess $N_\text{D}$ channels and each $m$ -th feature has dimensions $N_\text{D} \times \times H / 2^m \times W / 2^m$.
Subsequently, the backbone feature $\mat{F}^\text{bb}_1$ and pixel feature $\mat{F}^\text{px}_2$ are combined using a \textit{Feature Pyramid Network}~\cite{Lin2016Feature}, succeeded by task-specific 2-layer MLPs that produce features $\matF_\text{mask}$ and $\matF_\text{depth}$. 
The resulting \textit{task features} have dimensions $N_\text{D} \times H/2 \times W/2$.

\paragraph{Unified query decoder.} 
The unified decoder represents objects through \textit{shared queries} that embed the visual features of objects in the scene. 
These queries are refined in an iterative process~\cite{Cheng2022Masked}, and are ultimately used to predict the objects' segmentation and depth.
We initialize the queries $\matQ_0\in\R^{N_\text{Q} \times N_\text{D}}$ (the amount is $N_\text{Q}$) with learnable parameters $\matQ_\ell \sim \mathcal{N}(0, 1{\times}10^{-2})$, and iteratively refine them through a series of $N_\text{B}$ decoder blocks.
At each \mbox{$b$-th} decoder block, queries $\matQ_{b}$ are attended to pixel features $\matF^\text{px}_k$ through \textit{masked-attention}~\cite{Cheng2022Masked}, which allows queries to target specific localized regions of the pixel features.
One such iteration from $b-1$ to $b$ is given by
\begin{align}
    \hat{\matQ}_b &= \mathrm{MaskAttn}(\matQ_{b-1}, \matF^\text{px}_{k}, \matM_{b-1})
\text{,}\label{eqn:maskattn}
\end{align}
where $\matM_{b-1}$ is the mask generated at the previous layer, upsampled to match the dimensions of $\matF^\text{px}_k$.
This process starts from the lowest-resolution pixel feature ($k{=}P$) and decrementally progresses to the highest-resolution pixel feature ($k{=}2$), beyond which the iteration is reinitiated. 
This can be expressed as
\begin{align}
    k &= P- (b{-}1)~\mathrm{mod}~(P{-}1)
~\text{.}\label{eqn:iterator-k}
\end{align}
After each iteration, self-attention and a feedforward network are applied to the queries for updating, \ie
\begin{align}
    \matQ_{b} &= \mathrm{FFN}\left(\mathrm{SelfAttn}(\hat{\matQ}_{b})\right)
~\text{.}\label{eqn:selfattn-ffn}
\end{align}
Task-specific 3-layer MLPs generate mask kernels $\matK^\text{mask}_b$ and depth kernels $\matK^\text{depth}_b$ from the updated queries $\matQ_b$.
Subsequently, the segmentation mask $\matM_b$ and the normalized depth map $\hat{\matD}_b$ are predicted via
\begin{align}
    \matM_{b} &= \sigma(\matK^\text{mask}_b * \matF_\text{mask})
~\text{, and}\label{eqn:mask-pred}\\
    \hat{\matD}_{b} &= \sigma(\matK^\text{depth}_b * \matF_\text{depth})
~\text{,}\label{eqn:depth-pred}
\end{align}
where $*$ denotes a pointwise convolution operation, and $\sigma(\cdot)$ is the sigmoid function.
The next block further refines the updated queries using the masks, repeating the process until the final layer $b=N_{B}$ is reached.
The classification logits $\vec{\ell}_\text{class}$ are obtained by applying a learnable transform $f_\text{class}(\cdot)$ to the queries, expressed as
\begin{align}
    \vec{\ell}_\text{class} &= f_\text{class}(\matQ_{N_\text{B}}) \in \R^{N_\text{Q}\times N_\text{C}}
~\text{,}\label{eqn:class-pred}
\end{align}
where $N_\text{C}$ is the number of classes.

\paragraph{Panoptic segmentation.}
The panoptic merging algorithm from~\cite{Cheng2021MaskFormer} is utilized to process the mask predictions $\matM_b$ obtained from the final query decoder layer $b={N_\text{B}}$, thereby producing the panoptic segmentation output.

\paragraph{Object tracking.}
The tracking process operates through an association-based mechanism.
For frame $t$, let $\bar{\matQ}(t)$ denote the query subset representing detected objects.
The algorithm computes a pairwise cosine similarity matrix between queries $\bar{\matQ}(t)$ and $\bar{\matQ}(t-1)$, establishing an assignment cost matrix between objects in consecutive frames. 
The optimal object associations are then determined using the Jonker-Volgenant algorithm~\cite{Jonker1987shortest}, enabling the propagation of object identities from the previous frame to the current one.

\paragraph{Monocular depth.}
The normalized depth maps $\hat{\matD}\in\interval{0}{1}$ are transformed into metric depth values $ \matD \in \interval{d_\text{min}}{d_\text{max}}$ via min-max denormalization. 
This can be expressed as
\begin{align}
   \matD = r\hat{\matD} + \mu
~\text{,}\label{eqn:depth-scaleshift}
\end{align}
where $r$ and $\mu$ denote the scene's scale and shift parameters, respectively.
These parameters are derived as $r = d_\text{max} - d_\text{min}$ and $\mu = d_\text{min}$, where $\{d_\text{min},d_\text{max}\}$ are hyperparameters that define the depth range for a given dataset.
To generate the final depth map, each query-wise depth map is "copy and pasted" into the corresponding panoptic segment~\cite{Qiao2020ViP,Petrovai2023MonoDVPS,Yuan2021PolyphonicFormer,JiYeon2024}.

% ---------- MAIN CONTRIBUTIONS -----------
\subsection{Hybrid query decoder}
\label{sec:method-predictor}
\begin{figure}
    \centering
    % \begin{adjustbox}{max width=\linewidth}
    % \centering
    % \includestandalone[mode=buildnew]{figure_hybrid_block}
    % \end{adjustbox}
    \includegraphics[width=\linewidth]{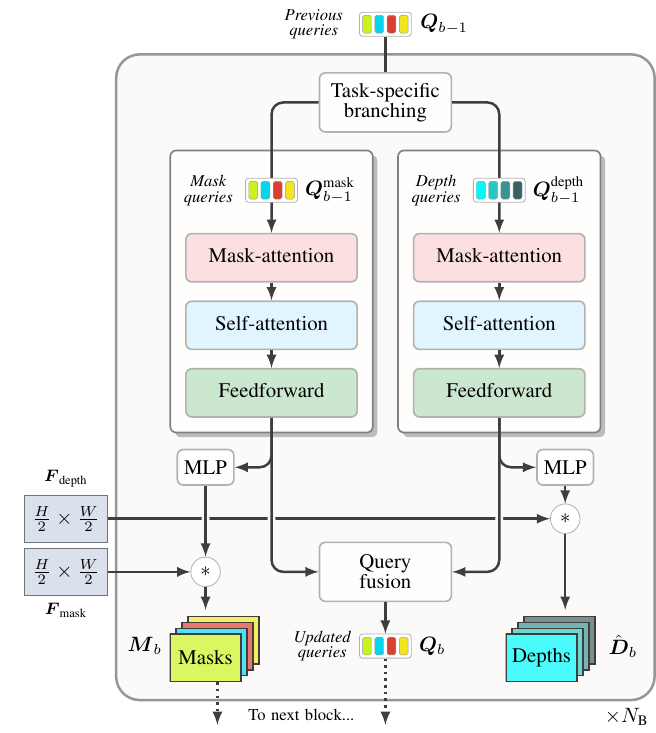}
    \caption{
        \textbf{Hybrid decoder block.} 
        Dedicated branches for each task are responsible for the processing and refinement of learnable queries $\matQ_{b-1}$.
        Subsequently, these refined task-specific queries are fused into a single shared query $\matQ_b$ at the interface between the different blocks.
    }
    \label{fig:predictor}
\end{figure}

We present a \textit{hybrid query decoder} that extends the \textit{unified query decoder} of the baseline network (\cref{sec:method-baseline}). 

% ~~~~~~~~~~ Hybrid queries ~~~~~~~~~~
\subsubsection{Hybrid decoder block}
\label{sec:method-multibranch-block}
The objective of this research is to identify a compromise between fully shared decoder architectures, \eg where all information about all tasks is encoded within a single query, versus conventional decoders that have specialized embeddings tailored for each task.
The proposed \textit{hybrid decoder block} effectively integrates the advantages of shared and task-specific representations through a branched design, as illustrated in \cref{fig:predictor}.

The motivation for adopting this hybrid approach stems from the observation that while shared representations offer efficiency and implicit multi-task learning, they may limit the model's ability to capture task-specific nuances. 
Conversely, fully separated representations allow for specialized learning but fail to capture potential synergies between tasks and are less efficient. 
The proposed hybrid query decoder aims to leverage the strengths of both paradigms.

At the core of the hybrid query decoder lies the concept of task-specific branching within each decoder layer, followed by a fusion into a shared representation at each decoder layers' interface. 
This design allows the model to learn task-specific features, while maintaining a shared representation that can benefit from cross-task information sharing. 
The process can be broken down into two main steps, as follows.

\paragraph{Task-specific branching}
Each $b$-th decoder block begins with shared queries $\matQ_{b-1}$ emanating from the preceding block. 
First, these queries are divided into task-specific queries $\matQ^\text{mask}_{b-1}$ and $\matQ^\text{depth}_{b-1}$ through a learnable linear transform.
Second, the task-specific queries are updated in separate branches through masked-attention \cref{eqn:maskattn}, followed by self-attention and feedforward layers \cref{eqn:selfattn-ffn}.
This yields updated queries $\matQ^\text{mask}_{b}$ and $\matQ^\text{depth}_{b}$ that have been attended to the (shared) pixel features $\matF^\text{px}_k$, whereby in the \textit{hybrid} scenario, task-specific nuances can be captured.

\paragraph{Query fusion.}
Updated queries $\matQ^\text{mask}_b$ and $\matQ^\text{depth}_b$ are fused into a shared query $\matQ_{b}$.
To this end, a learnable linear transformation $f_\text{fuse}(\cdot)$ is utilized, followed by an addition operation, leading to the expression  
\begin{align}
    \matQ_{b} = \textrm{norm}_2\left(f_\text{fuse}^\text{mask}(\matQ^\text{mask}_b) + f_\text{fuse}^\text{depth}(\matQ^\text{depth}_b) \right)
~\text{.}
\end{align}
The fused representation $\matQ_{b}$ undergoes L2 normalization to ensure stable training.

This hybrid approach offers several advantages. 
Primarily, it facilitates task-specific learning within each decoder layer, thus capturing subtle distinctions that might otherwise be overlooked in a completely unified approach.
Furthermore, query fusion at each layer interface facilitates the exchange of information between tasks, which may enhance overall performance and activate the inherent multi-task learning potential of shared representations~\cite{JiYeon2024} in the blocks that follow.

% ~~~~~~~~~~ Context adapter ~~~~~~~~~~
\subsubsection{Context adapter}
\label{sec:method-context}
\begin{figure}
    \centering
    % \begin{adjustbox}{max width=\linewidth}
    % \centering
    %\includestandalone[mode=buildnew]{figure_context_adapter}
    % \end{adjustbox}
    \includegraphics[width=\linewidth]{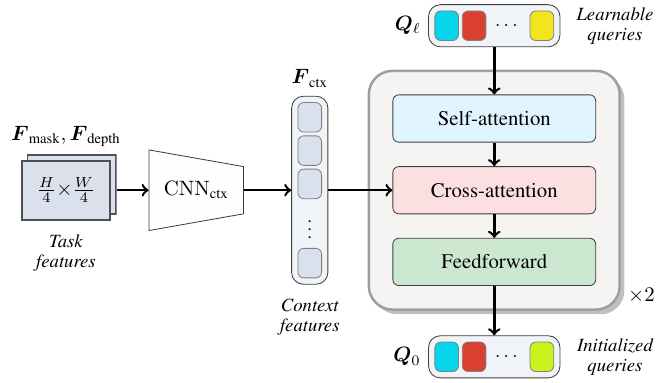}
    \caption{
        \textbf{Context adapter.} 
        The context feature $\matF_\text{ctx}$ serves as a condensed embedding of the task-specific features $\matF_\text{mask}$ and $\matF_\text{depth}$. 
        Learnable queries $\matQ_\ell$ undergo adaptation to the context feature $\matF_\text{ctx}$ via an attention network, producing initial queries $\matQ_0$.
    }
    \label{fig:ctx-adapter}
\end{figure}
The context adapter serves as an initial conditioning mechanism for the learnable queries $\matQ_\ell$.
This module has the primary purpose of seeding the initial queries $\matQ_0$, \ie before entering the decoder blocks, with a representation that has been adapted to the task features (see top-right of \cref{fig:architecture}).
Conceptually, this process can be viewed as the inverse of the hybrid query decoder principle: instead of aligning task-specific queries $\{\matQ^\text{depth}, \matQ^\text{mask}\}$ with shared pixel features $\matF^\text{px}$ (see \cref{sec:method-multibranch-block}), the learnable (shared) queries $\matQ_\ell$ are aligned with task-specific features $\{\matF_\text{depth},\matF_\text{mask}\}$, resulting in the generation of the initial queries $\matQ_0$.

Based on a 2-layer transformer decoder~\cite{Cheng2021MaskFormer}, the adapter uses cross-attention between learnable queries $\matQ_\ell$ and a context feature $\matF_\text{ctx}$, as depicted in \cref{fig:ctx-adapter}.
This context feature is derived from the concatenated task features via
\begin{align}
    \matF_\text{ctx} = \mathrm{CNN}_\text{ctx}\left(\left[\matF_\text{depth} ~~~ \matF_\text{mask}\right]\right)
~\text{,}
\end{align}
where $\mathrm{CNN}_\mathtext{ctx}(\cdot)$ denotes a convolutional block that serves to reduce dimensionality while effectively propagating information relevant to query initialization.

% ---------- OTHER CONTRIBUTIONS ----------
\subsection{Architectural improvements}
\label{sec:method-improvements}

The following straightforward improvements are proposed to the baseline network to improve its performance.

% ~~~~~~~~~~ Deep supervision ~~~~~~~~~~~
\subsubsection{Deep supervision}

In the \textit{Mask2Former}~\cite{Cheng2022Masked} architecture, the masks corresponding to
each query are utilized to progressively refine the localized regions to which queries are tuned. 
Since this yields mask predictions at the interfaces between decoder blocks, the mask losses can be applied directly to these intermediate masks. 
This process is known as \textit{deep supervision} and has been shown to improve network convergence as well as segmentation quality~\cite{Cheng2022Masked}.
Despite the absence of depth in the query refinement process, an analogous methodology can be implemented for the depth-estimation task.
This is accomplished simply by calculating the depth maps $\matD_b$ at each layer $b \in \{1,\cdots,N_\text{B}\}$ throughout the training phase, as opposed to merely generating the final depth map $\matD_{N_\text{B}}$, thereby facilitating the application of depth losses to this intermediate prediction. 
During inference, solely the final decoder layer produces depth estimations.

% ~~~~~~~~~~ Depth estimation ~~~~~~~~~~
\subsubsection{Depth estimation}
\label{sec:methods-improvements-depth}
We propose three enhancements to the depth estimation process.
These modifications result in increased training stability and improved depth-estimation performance, as demonstrated by the experimental results (\cref{sec:experiments-main}).
The enhancements are as follows.

\paragraph{Scale and shift.}
The proposed model effectively obviates the requirement for hyperparameters $\{d_\text{min},d_\text{max}\}$ by concurrently estimating the scale $r$ and shift $\mu$ parameters from the input data. 
To facilitate this, pixel feature $\matF^\text{px}_2$ undergoes a 2-layer CNN succeeded by a linear transformation. 
Exponential activation is applied to the scale parameter such that $0 \leq r < \infty$, while the shift parameter $\mu \in \R$ remains unconstrained.

\paragraph{Log-depth modeling.}
The sigmoid activation $\sigma(\cdot)$ is eliminated from \cref{eqn:depth-pred}, and the result is reinterpreted to predict unnormalized log-depth values directly, \ie \cref{eqn:depth-pred} is replaced by
\begin{align}
    \hat{\matD}_b &= \matK^\text{depth}_b * \matF_\text{depth} 
\text{.}\label{eqn:unnorm-log-depth}
\end{align}
Let $\vecD_q \in \R^{1 \times H \times W}$ and $\vecQ_q \in {1\times N_\text{D}}$ be elements that correspond to the $q$-th query in $\matD$ and $\matQ$, respectively. 
The query-wise normalized depths $\vecD^\text{norm}_q$ are then derived from \cref{eqn:unnorm-log-depth} via
\begin{align}
    \vecD^\text{norm}_q &= \frac{\hat{\vecD_q}- \mathrm{mean}(\hat{\vecD_q})}{\mathrm{std}(\hat{\vecD_q})} \gamma(\vecQ_q) + \beta(\vecQ_q)
\text{,}
\end{align}
where $\gamma(\cdot)$ and $\beta(\cdot)$ are learnable transforms that represent query-wise affine parameters.
Subsequently, the metric depths are computed, replacing \cref{eqn:depth-scaleshift} with 
\begin{align}
    \vecD_q &= r \left( \exp{\vecD^\text{norm}_q} + \mu \right)
~\text{,}
\end{align}
such that $\matD = \left[ \vecD_q \right]^{q=1}_{N_\text{Q}}$.

\paragraph{Dynamic depth merging.}
The current common practice in DVPS is to "copy and paste" each query-wise depth map into the corresponding panoptic segmentation masks~\cite{Yuan2021PolyphonicFormer,Petrovai2023MonoDVPS,JiYeon2024}. 
This leads to a final depth map that is highly sensitive to the quality of those masks.
To mitigate this effect, a dynamic merging algorithm is introduced.
First, the softmax scores $\vecS \in \R^{N_\text{Q}}$ are computed from classification logits $\vec{\ell}\in\R^{N_\text{Q}\times N_\text{C}}$ via
\begin{align}
    \vecS &= \max\left(~ \mathrm{softmax}(\vec{\ell}) ~\right)
    \text{.}
\end{align}
Next, the low-confidence depth estimates are discarded, and the scores $\vecS$ are used
to compute pixel-wise weights in the unity interval, specified by
\begin{align}
\matW &= \mathrm{softmax}(\frac{\vecS\Tr\matM}{\tau}) 
\end{align}
where temperature parameter $\tau$ controls the sharpness of the softmax.
Finally, the weighted average of $\matD\in \R^{N_\text{Q}\times H\times W}$ is computed pixel-wise using
weights $\matW\in [0,1]^{N_\text{Q}\times H\times W}$, resulting in the final depth map.

\subsection{Training and losses}
\label{sec:method-training}
\def\loss{\mathcal{L}}%
The composite loss function is defined as
\begin{align}
    \loss_\text{total} = 
        \lambda_\mathtext{mask}\loss_\mathtext{mask} + 
        \lambda_\mathtext{class}\loss_\mathtext{class} +
        \lambda_\mathtext{depth}\loss_\mathtext{depth} 
    \text{.}
\end{align}
The mask and classification components follow \textit{Mask2Former}~\cite{Cheng2022Masked}, utilizing the binary cross-entropy and DICE metric for $\loss_\text{mask}$ with $\lambda_\text{mask}=5$, and employing the cross-entropy loss for $\loss_\text{class}$ with $\lambda_\text{class}=1$.
The depth loss $\loss_\text{class}$ is defined as the sum of the scale-invariant logarithmic loss~\cite{Eigen2014Depth} and root mean-squared error, with $\lambda_\text{depth}=1$.

% ========== E X P E R I M E N T S ===========
\section{Experiments}
\label{sec:experiments}
% TABLE: Main results on all tasks
\begin{table*}[t]

\centering
\begin{threeparttable}
\begin{tabularx}{\linewidth}{%
@{}l@{\hskip2em}X@{\hskip3em}c@{\quad}c@{\quad}c@{\hskip3em}c@{\quad}c@{}%
}
%------------------------------------------------------------
\toprule                                                    %
% -- HEADER ------------------------------------------------- 
\multirow{2}{*}{\thead{Method}}          & 
\multirow{2}{*}{\thead{Backbone}}       & 
\multicolumn{3}{@{}c@{\hskip3em}}{\thead{Depth-aware video panoptic}}  &
\multicolumn{2}{@{}c@{}}{\thead{Monocular depth}}  \\
% -----------------------------------------------------------
% \cmidrule{3-5@{~}}
% \cmidrule{@{~}6-7}
& & %     
\small{{DVPQ}$_\text{All}$$\uparrow$}  &
\small{{DVPQ}$_\text{Thing}$$\uparrow$}  &
\small{{DVPQ}$_\text{Stuff}$$\uparrow$}  &
\small{{Abs.Rel.}~$\downarrow$}             &  
\small{{RMSE}~$\downarrow$}  \\ 
% ----------------------------------------------------------- %
\midrule                                                      %
% -- DATA --------------------------------------------------- %
ViP-DeepLab~\cite{Qiao2020ViP} & ResNet-50 &            %             
\num{42.0}  & \num{27.6}    & \num{51.5}    &                 % DVPQ
\num{0.070} & \num{3.67}    \\%& $<1$     \\               % AbsRel, RMSE, FPS
% ----------------------------------------------------------- %
MonoDVPS~\cite{Petrovai2023MonoDVPS} & ResNet-50        &                 
\num{48.8}  & \num{31.0}    & \num{61.7}    &                 % DVPQ
\num{0.070} & \num{3.67}   \\% & \num{10.0}     \\               % AbsRel, RMSE, FPS
% ----------------------------------------------------------- %
PolyphonicFormer~\cite{Yuan2021PolyphonicFormer}  & ResNet-50 & 
\num{48.1}  & \num{35.6}    & \num{57.1}    &                 % DVPQ
\num{0.081} & \num{4.01}   \\% & \num{1.4}     \\                % AbsRel, RMSE, FPS
% ----------------------------------------------------------- %
UniDVPS~\cite{JiYeon2024} & ResNet-50 &                    % Method, Name
\num{51.8}  & \num{37.1}    & \num{62.5}    &                 %  % DVPQ
\num{0.067} & \num{3.88}   \\% &  \num{9.2}    \\                % AbsRel, RMSE, FPS
% ----------------------------------------------------------- %
% \midrule
% ----------------------------------------------------------- %
\textbf{Multiformer} ({ours})         & ResNet-50        &                 % 
\textbf{\num{54.8}}  & \textbf{\num{37.4}}    & \textbf{\num{67.4}}    &                 % DVPQ
\textbf{\num{0.066} }& \textbf{\num{3.25}}  \\% & \num{6.5}     \\                % AbsRel, RMSE, FPS
% ----------------------------------------------------------- %
\midrule % 
% ----------------------------------- ll--------------------- %
PolyphonicFormer~\cite{Yuan2021PolyphonicFormer}  & Swin-B & 
\num{55.4}  & \num{43.3}    & \num{63.6}    &                 % DVPQ
\num{0.065} & \num{3.8}   \\%& --     \\                % AbsRel, RMSE, FPS
% ----------------------------------------------------------- %
% \cmidrule{3-7}
% ----------------------------------------------------------- %
% 
\textbf{Multiformer} ({ours})       & Swin-B     &                     % 
\textbf{\num{59.4} }& \textbf{\num{46.0}}  & \textbf{\num{69.2}}   &                    % DVPQ
\textbf{\num{0.048}} &\textbf{ \num{2.81}} \\%& \num{4.1}      \\                 % AbsRel, RMSE, FPS
% -----------------------------------------------------------
%\ours*     & $\dots$       &                                 % Method, Name
%$\dots     & $\dots$       & $\dots$       &                 % DVPQ
%$\dots     & $\dots$       & $\dots$       & $\dots$       \\% AbsRel, RMSE, FPS
%------------------------------------------------------------
\bottomrule
%------------------------------------------------------------
\end{tabularx}
\end{threeparttable}

\caption{
    \textbf{Main results.} 
    Comparison of depth-aware video panoptic segmentation and depth estimation performance on Cityscapes-DVPS.
}
\label{tab:results-main}
\end{table*}
% ---------- DATASETS ----------
\subsection{Datasets}
\label{sec:experiments-datasets}
\noindent
\textbf{Cityscapes-DVPS}~\cite{Qiao2020ViP} is the de-facto standard dataset for evaluating the DVPS task, extending the Cityscapes-VPS~\cite{Kim2020Video} dataset with depth annotations.
The dataset consists of 450 videos, wherein each 30-frame video has 6 annotated frames (5 frames between annotations).
The training and validation sets have 2,400~and 300~annotated frames, respectively.
There are 19 classes (8 `thing' and 11 `stuff') in the dataset, following the Cityscapes \cite{Cordts2016Cityscapes} labeling scheme.

\noindent
\textbf{SemKITTI-DVPS}~\cite{Qiao2020ViP} is derived from the odometry split of the \textit{KITTI}~\cite{Geiger2012Are} dataset.
The dataset comprises 11 videos of varying lengths that are divided into 10 training videos (19,130~frames) and 1 validation video (4,071~frames).
All frames possess sparse semantic annotations acquired by projecting panoptic-labeled 3D point clouds from 
\textit{SemanticKITTI}~\cite{Behley2019Semantickitti} onto the image plane. 
This dataset includes 19 classes (8 `thing' and 11 `stuff').

% ----------- METRICS ----------
\subsection{Metrics}
\label{sec:experiments-metrics}
The results are presented using their canonical evaluation metrics, as enumerated below.
\begin{itemize}
\item \textbf{Overall performance}, \ie depth-aware video panoptic segmentation images, are assessed using Depth-aware Video Panoptic Quality (DVPQ)~\cite{Qiao2020ViP}.
\item \textbf{Panoptic segmentation} is evaluated using {Panoptic Quality (PQ)}~\cite{Kirillov2018Panoptic} and {Video Panoptic Quality (VPQ)}~\cite{Kim2020Video}. 
\item \textbf{Monocular depth estimation} accuracy is quantified via the {Absolute Relative Error (AbsRel)} and {Root Mean-Squared Error (RMSE)}~\cite{Eigen2014Depth}. 
\end{itemize}

% ---------- IMPLEMENTATION DETAILS ----------
\subsection{Implementation details}
\label{sec:experiments-implementation}
The proposed models are implemented in PyTorch. 
\textit{ResNet} \cite{He2015Deep} and \textit{SwinTransformer} \cite{Liu2021Swin} are adopted as backbone networks, initialized using weights pre-trained for ImageNet classification.
Unlike some approaches, the \textit{Multiformer} does not apply test-time augmentation (TTA)~\cite{Qiao2020ViP, Petrovai2023MonoDVPS, Chen2020Naive} or additional pre-training~\cite{Qiao2020ViP, Petrovai2023MonoDVPS, Yuan2021PolyphonicFormer, JiYeon2024}. 
The model is trained for 20K steps on 4 \textit{NVIDIA H100}-GPUs using the \textit{AdamW} optimizer at \num{5e-4} learning rate, following \textit{Mask2Former}~\cite{Cheng2022Masked} settings unless otherwise specified.

% ---------- MAIN RESULTS ----------
\subsection{Main results}
\label{sec:experiments-main}

The \textit{Multiformer} demonstrates strong performance for depth-aware video panoptic segmentation (DVPS) and monocular depth estimation. 
\cref{tab:results-main} presents a comprehensive comparison of our method with state-of-the-art approaches on the Cityscapes-DVPS dataset.
With the ResNet-50~\cite{He2015Deep} backbone, the proposed method outperforms UniDVPS~\cite{JiYeon2024} by \num{3.0} DVPQ (\textit{all}) points, while also improving depth estimation accuracy.
When using the more powerful Swin-B~\cite{Liu2021Swin} backbone, the Multiformer surpasses PolyphonicFormer~\cite{Yuan2021PolyphonicFormer} by \num{4.0} DVPQ (\textit{all}) points.

% TABLE: Long-term DVPS results
\begin{table}[!t]
    \centering
    
\begin{adjustbox}{max width=\linewidth}
\begin{tabularx}{\linewidth}{@{}X@{~}r@{~}|@{~~}c@{~~}c@{~~}c@{~~}c@{~~}c@{}}
\toprule %-----------------------------------------------------------
%\multicolumn{2}{@{}l}{\small\it SemanticKITTI-DVPS} &%\\
\multicolumn{2}{@{}l}{\thead{DVPQ}$_\kappa^\lambda$} &%\\
%\multicolumn{2}{@{}l}{~} &
\footnotesize $\kappa{=}1$  & \footnotesize $\kappa{=}5$& \footnotesize $\kappa{=}10$& \footnotesize $\kappa{=}20$ & \footnotesize Avg. \\
\midrule
\multicolumn{2}{@{}l}{ViP-DeepLab WR-50~\cite{Zagoruyko2016Wide}}
                       & \num{48.9} & \num{45.8} & \num{44.4} & \num{43.4} &  \num{45.6}  \\
\midrule
\multirow{4}{*}{
    \small \makecell[l]{PolyphonicFormer~\cite{Yuan2021PolyphonicFormer}\\Swin-B}
}
& \footnotesize $\lambda{=}0.50$ & \num{58.5} & \num{52.0} & \num{50.6} & \num{49.9} & \num{52.8} \\
& \footnotesize $\lambda{=}0.25$ & \num{56.3} & \num{49.7} & \num{48.4} & \num{47.7} & \num{50.5} \\
& \footnotesize $\lambda{=}0.10$ & \num{41.8} & \num{35.1} & \num{33.7} & \num{33.0} & \num{35.9} \\
& \footnotesize Avg.             & \num{52.2} & \num{45.6} & \num{44.2} & \num{43.4} & \num{46.4} \\
\cmidrule{2-7}
\multirow{4}{*}{
    \small \makecell[l]{\textbf{Multiformer} (ours)\\Swin-B}
}
& \footnotesize $\lambda{=}0.50$ & \num{56.6}  & \num{55.2} & \num{54.6} & \num{49.6} & \num{54.0}  \\
& \footnotesize $\lambda{=}0.25$ & \num{51.4}  & \num{49.6} & \num{49.5} & \num{48.5} & \num{49.7}  \\
& \footnotesize $\lambda{=}0.10$ & \num{49.1}  & \num{47.2} & \num{46.6} & \num{46.2} & \num{47.3}  \\
& \footnotesize Avg.             & \bf\num{52.3}  & \bf\num{50.6} & \bf\num{50.2} & \bf\num{48.2} & \bf\num{50.3}  \\
\bottomrule % --------------------------------------------------------
\end{tabularx}
\end{adjustbox}

    \caption{
        \textbf{DVPQ scores} for different window size ($\kappa$) and relative error threshold ($\lambda$) on SemKITTI-DVPS.
    }
    \label{tab:result-combined}
\end{table}
The DVPQ-metric is evaluated in varying temporal window sizes and depth thresholds, as shown in \cref{tab:result-combined}. 
The \textit{Multiformer} demonstrates improved average DVPQ performance and is robust across various temporal window sizes ($\kappa$) and depth thresholds ($\lambda$). 
The proposed method maintains high performance even with larger temporal windows and stricter depth thresholds, outperforming PolyphonicFormer~\cite{Yuan2021PolyphonicFormer} in multiple settings.

\subsection{Ablation studies}
\paragraph{Depth estimation.}
% TABLE: Depth estimation ablations
\begin{table}[!t]
    \centering
    
\centering
\begin{tabularx}{\linewidth}{@{}X c c@{}}
\toprule
% -- HEADER --------------
\thead{Method} & \thead{Abs.Rel. $\bm{\downarrow}$} & \thead{RMSE $\bm{\downarrow}$} \\
% --------------------------
\midrule
% -----------------------------
PanopticDepth~\cite{Gao2022PanopticDepth}   & --          & \num{6.91} \\
MonoDVPS~S-MDE\cite{Petrovai2023MonoDVPS}   & \num{0.082} & \num{4.91} \\
MonoDVPS~\cite{Petrovai2023MonoDVPS}        & \num{0.070} & \num{3.67} \\
UniDVPS~\cite{JiYeon2024}                      & \num{0.067} & \num{3.88} \\
% ------------------------------
\midrule
% ----------------------------
Multiformer w/ minmax (\ref{eqn:depth-scaleshift})            & \num{0.069} & \num{3.54} \\
~${-}$~dynamic merge                       & \num{0.073} & \num{3.74} \\
~~${-}$~context adapter                     & \num{0.074} & \num{3.84} \\ 
~~~${-}$~scale/shift estimator  & \num{0.078} & \num{3.89} \\
~~~~${-}$~deep supervision                    & \num{0.085} & \num{4.64} \\ 
% ------------------------------
\midrule%\cmidrule{2-3}
% ----------------------------
\textbf{Multiformer} (ours)                           & \textbf{\num{0.066}} & \textbf{\num{3.35}} \\
~${-}$~dynamic merge                       & \num{0.076} & \num{3.81} \\
~~${-}$~context adapter                     & \num{0.078} & \num{3.86} \\
~~~${-}$~query-wise affine                   & \num{0.085} & \num{4.39} \\
~~~~${-}$~deep supervision                    & \num{0.091} & \num{4.65} \\
% -----------------------------
\bottomrule
\end{tabularx}

    \caption{\textbf{Monocular depth estimation.} Evaluated on Cityscapes-DVPS using $N_\text{B}=9$ decoder blocks (L) and a ResNet-50 backbone}
    \label{tab:result_mono_depth}
\end{table}
The proposed depth estimation improvements (see \cref{sec:methods-improvements-depth}) are experimentally validated by ablation, as summarized in \cref{tab:result_mono_depth}.
The improved \textit{Multiformer} model achieves comparable performance in depth estimation compared to previous segmentation-guided methods. 
The removal of dynamic merging, context adapter, query-wise affine transformation, and deep supervision all lead to performance degradation.

\paragraph{Scaling properties.}
% TABLE: Scaling amount of layers
\begin{table}[!t]
    
\begin{tabularx}{\linewidth}{@{}X c c c@{}}
\toprule
\thead{Variant} & \thead{$\bm{N}_\text{B}$} & \thead{DVPQ $\bm{\uparrow}$} & \thead{$\bm{N}_\text{P}$}  \\
\midrule
Multiformer-S   & 3      & \num{52.7}   & \num{18}~M \\
Multiformer-M   & 6      & \num{53.2}   & \num{25}~M \\
Multiformer-L   & 9      & \num{54.8}   & \num{32}~M \\
\bottomrule
\end{tabularx}

    \caption{
        \textbf{Model variants.} 
        DVPQ and number of parameters $N_\text{P}$ under varying number of query decoder blocks $N_\text{B}$. Evaluated on Cityscapes-DVPS.
    }
    \label{tab:variants}
\end{table}
The impact of scaling the proposed model is investigated by modulating the number of query decoder blocks $N_\text{B}$, as shown in \cref{tab:variants}.
For the remaining experiments, the Multiformer-S model is adopted, which has $N_\text{B}=3$ query decoder blocks.

% ~~~~~~~~~~ Query decoder ablations ~~~~~~~~~~
\paragraph{Query decoder design.}
% FIGURE: Decoder architecture definitions
\begin{figure*}[!thb]
    \centering
    \begin{subfigure}{.195\linewidth}%
        % \begin{adjustbox}{max width=\linewidth}
        %     \includestandalone[mode=buildnew,page=1]{figure_decoder_variant}
        % \end{adjustbox}
        \includegraphics[width=\linewidth, page=1]{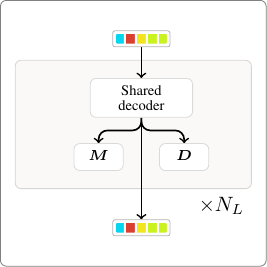}
        \caption{Unified ~\cite{JiYeon2024}}
    \end{subfigure} 
    \hfill
    \begin{subfigure}{.195\linewidth}
        % \begin{adjustbox}{max width=\linewidth}
            % \includestandalone[mode=buildnew,page=2]{figure_decoder_variant}
        % \end{adjustbox}
        \includegraphics[width=\linewidth, page=2]{figures/figure_decoder_variant.pdf}
        \caption{Parallel}
    \end{subfigure} 
    \hfill
    \begin{subfigure}{.195\linewidth}
        % \begin{adjustbox}{max width=\linewidth}
        %     \includestandalone[mode=buildnew,page=3]{figure_decoder_variant}
        % \end{adjustbox}
        \includegraphics[width=\linewidth, page=3]{figures/figure_decoder_variant.pdf}
        \caption{Concat}
    \end{subfigure} 
    \hfill
    \begin{subfigure}{.195\linewidth}
        % \begin{adjustbox}{max width=\linewidth}
        %     \includestandalone[mode=buildnew,page=4]{figure_decoder_variant}
        % \end{adjustbox}
        \includegraphics[width=\linewidth, page=4]{figures/figure_decoder_variant.pdf}
        \caption{Sequential}
    \end{subfigure} 
    \hfill
    \begin{subfigure}{.195\linewidth}
        % \begin{adjustbox}{max width=\linewidth}
        %     \includestandalone[mode=buildnew,page=5]{figure_decoder_variant}
        % \end{adjustbox}
        \includegraphics[width=\linewidth, page=5]{figures/figure_decoder_variant.pdf}
        \caption{Hybrid (\textit{ours})}\label{fig:decoder(hybrid)}
    \end{subfigure} 
    \caption{
        \textbf{Design space exploration.} 
        Each diagram shows a variant of the \textit{query decoder block} design (\cref{sec:method-baseline}), where \textit{shared} or \textit{task-specific} queries are used to predict masks $\matM$ and depths $\matD$.
        Left to right:
        (a) uses shared queries and a shared decoder;
        (b) uses task-specific queries and decoders; 
        (c) uses a shared decoder on channel-wise concatenated task-specific queries; 
        (d) uses fuses task-specific queries between sequential task-specific decoders;
        (e) uses task-specific decoders that subsequently fuse into shared queries (see \cref{sec:method-multibranch-block}).
    }
    \label{fig:decoder}
\end{figure*}

Variations on the query decoder design (see \cref{fig:decoder}) are explored and evaluated. 
The results of this design space exploration are presented in \cref{tab:decoder}.
The hybrid query decoder block (\cref{fig:decoder(hybrid)}) outperforms the other designs, demonstrating the benefit of the proposed hybrid design principles.

% ~~~~~~~~~~ Baseline reproduction ~~~~~~~~~~~
\paragraph{Component analysis.}

To wrap up the experiments, results of building the experimental setup from the baseline (\cref{sec:method-baseline}) to the final improved Multiformer are summarized in \cref{tab:baseline}.
First, \textit{Mask2Former}~\cite{Cheng2022Masked} is adapted to the depth-aware video panoptic segmentation task, reproducing the methods proposed in \textit{UniDVPS}~\cite{JiYeon2024}. 
The results show that the reproduced baseline (UniDVPS-M2F) performs approximately on par with \textit{UniDVPS}~\cite{JiYeon2024}.
However, a slight performance degradation is observed, likely due to lack of pre-training.
Subsequently, the proposed baseline is upgraded with the hybrid decoder block, context adapter, and the improvements discussed in \cref{sec:method-improvements}.
Finally, the components \textit{hybrid decoder block} and \textit{context adapter} are systematically excluded to show the degradation associated with each individual element.
The analyses indicate that the hybrid decoder block exerts a significant influence on performance, with potential enhancements achievable through the incorporation of the context adapter.
% TABLE: Decoder ablation results
\begin{table}
    
\centering
\begin{threeparttable}
% == TABLE ENVIRONMENT ===================================================================================
\begin{tabularx}{\linewidth}{@{}X c l c@{}}
\toprule %------------------------------------------------------------------------------------------------
\thead{Method}   &  \thead{$\bm{N}_\text{B}$}   & \thead{Decoder block} & \thead{DVPQ~$\bm{\uparrow}$} \\
\midrule % -----------------------------------------------------------------------------------------------
UniDVPS~\cite{JiYeon2024} & 3 & & \num{50.3} \\
UniDVPS-M2F~\tnote{(i)} & 3 & & \num{50.0} \\
% -----------------------------------------------------
\midrule
Multiformer   & 3 & (a) Unified     & \num{52.3} \\
Multiformer   & 3 & (b) Naive             & \num{45.7} \\
Multiformer   & 3 & (c) Concat      & \num{51.2} \\
Multiformer   & 3 & (d) Sequential        & \num{52.4} \\
\textbf{Multiformer} (ours)   & 3 & (e) \textbf{Hybrid}            & \textbf{\num{52.7}} \\
\cmidrule{2-4}
\textbf{Multiformer} (ours)   & 9 & (e) \textbf{Hybrid}            & \textbf{\num{54.8}} \\
% -----------------------------------------------------
\bottomrule 
\end{tabularx}
% == TABLE FOOTNOTES =====================================================================================
\begin{tablenotes}
    \item[(i)] our Mask2Former-based~\cite{Cheng2022Masked} reproduction.
\end{tablenotes}
% ========================================================================================================
\end{threeparttable}

    \caption{
        \textbf{Decoder architectures.} 
        Evaluated on Cityscapes-DVPS using ResNet-50 as the backbone.
        The decoder designs are depicted in \cref{fig:decoder}, and the number of decoder blocks is $N_\text{B}$.
    }
    \label{tab:decoder}
\end{table}
% TABLE: Baseline network
\begin{table}
    
\begin{threeparttable}
\begin{tabularx}{\linewidth}{@{}X@{}c@{~~}c@{~~}c@{~}c@{}}
\toprule
\thead{Method} & 
\thead{PQ$\bm{\uparrow}$} & 
\thead{VPQ$\bm{\uparrow}$} & 
\thead{DVPQ$\bm{\uparrow}$} & 
\thead{$\bm{N}_\text{P}$} \\
\midrule
%\MaskTFormer*         & --   & --     & --    &  --    \\
%\midrule
\multicolumn{5}{c}{\small\textit{Baseline model}} \\
UniDVPS~\cite{JiYeon2024}       & 65.0      & --        & 50.3      & 11.5~M \\
\midrule
\multicolumn{5}{c}{\small\textit{Reproduced model}} \\
Mask2Former\tnote{(i)}~\cite{Cheng2022Masked}
                                & 63.9      & --        & --        & 15.8~M \\
\ensuremath{+}~query tracker    & 63.4      & 56.5      & --        & 15.8~M  \\
\ensuremath{+}~depth            & 63.8      & 55.8      & 49.8      & 16.2~M  \\
UniDVPS-M2F\tnote{(ii)}          & 63.9      & 56.1      & 50.0      & 11.8~M \\
\midrule
\multicolumn{5}{c}{\small\textit{Proposed model}} \\
\textbf{Multiformer} (ours)      & \bf 65.2  & \bf 57.5  & \bf 52.7  & 18.0~M    \\ 
${-}$~\mbox{context adapter}         & 65.1      & 57.1      & 52.3      & 15.9~M   \\ % min 2.1M
${-}$~\mbox{hybrid decoder block}      & 64.9      & 56.6      & 50.2      & 12.8~M \\
\bottomrule
\end{tabularx}
\begin{tablenotes}
\item[(i)] based on publicly available implementation~\cite{Cheng2022Masked}.
\item[(ii)] align the number of parameters $N_\text{P}$ with~\cite{JiYeon2024} .
\end{tablenotes}
\end{threeparttable}

    \caption{
        \textbf{Baseline evaluation.} 
        Evaluated on Cityscapes-DVPS using ResNet-50 as the backbone and $N_\text{B}=3$ decoder blocks (S).
    }
    \label{tab:baseline}
\end{table}
% ========== C O N C L U S I O N ==========
\section{Conclusion}
\label{sec:conclusion}

We have introduced \textit{Multiformer}, a novel depth-aware video panoptic segmentation approach exploring the balance of shared and task-specific object representations. 
The proposed model leverages the concept of a hybrid query decoder in multi-task visual understanding, where tasks can be of different nature.
Key innovations include a hybrid decoder block with task-specific attention mechanisms for depth estimation and segmentation, capturing the nuances of each task.
The resulting task representations are fused at the interface between the decoder blocks, allowing cross-task interaction.
Experimental findings show that the proposed model outperforms existing methods in standard benchmarks, achieving improved performance in depth-aware video panoptic segmentation and its component tasks. 
Future work could explore the benefit of the proposed hybrid approach in other multi-task vision problems, as well as investigate ways to further improve the efficiency and scalability of the model. 

\section*{Acknowledgments}
\label{sec:acknowledgements}
The author expresses gratitude to Prof. P.H.N. De~With and Dr. F. van~der~Sommen for their thorough review and experimental corroboration of the results.
% Acknowledgement to funding partner NWO (NEON project)
This publication is part of the \textit{NEON} project with file number 17628 of the \textit{Crossover} research program, which is (partly) financed by the Dutch Research Council (NWO). 
% Acknowledgement to SURF for providing access to Snellius
The Dutch national compute infrastructure was used with the support of the SURF Cooperative using grant EINF-5438.
% ========== R E F E R E N C E S ==========
{\small
\bibliographystyle{ieee_fullname}
\bibliography{00ms}
}

% =========================================
\end{document}